\documentclass[runningheads]{llncs}
\usepackage[T1]{fontenc}
\usepackage{graphicx}
\usepackage{fancyhdr}
\usepackage{changepage} 
\usepackage{listings} 
\usepackage{subfigure}
\usepackage{hyperref}
\usepackage{booktabs}
\usepackage{multirow} 
 
\usepackage{graphicx}
\begin{document}
\title{Deep Learning-based Compression Detection for explainable Face Image Quality Assessment}
\titlerunning{Deep Learning-based Compression Detection}
%
\author{Laurin Jonientz\inst{1,2}, Johannes Merkle\inst{2},  Christian Rathgeb\inst{2,3}, Benjamin Tams\inst{2} and Georg Merz\inst{1}}
\authorrunning{L. Jonientz et al.}
%
\institute{$^1$Technische Hochschule Brandenburg, Brandenburg an der Havel, Germany\\
$^2$secunet Security Networks AG, Essen, Germany\\
$^3$Hochschule Darmstadt, Darmstadt, Germany\\ 
\email{\{laurin.jonientz,johannes.merkle\}@secunet.com}}
\maketitle              
\begin{abstract}
    The assessment of face image quality is crucial to ensure reliable face recognition. In order to provide data subjects and operators with explainable and actionable feedback regarding  captured face images, relevant quality components have to be measured. Quality components that are known to negatively impact the utility of face images include JPEG and JPEG 2000 compression artefacts, among others. Compression can result in a loss of important image details which may impair the recognition performance. In this work, deep neural networks are trained to detect the  compression artefacts in a face images. For this purpose, artefact-free facial images are compressed with the JPEG and JPEG 2000 compression algorithms. Subsequently, the PSNR and SSIM metrics are employed to obtain training labels based on which neural networks are trained using a single network to detect JPEG and JPEG 2000 artefacts, respectively. The evaluation of the proposed method shows promising results: in terms of detection accuracy, error rates of 2-3\% are obtained for utilizing PSNR labels during training. In addition, we show that error rates of different open-source and commercial face recognition systems can be significantly reduced by discarding face images exhibiting severe compression artefacts. To minimize resource consumption, \mbox{EfficientNetV2} serves as basis for the presented algorithm, which is available as part of the OFIQ software. 

\end{abstract}

\section{Introduction}\label{sec:intro}

Biometrics have become increasingly important in various application scenarios including transnational deployments such as the Entry-Exit System (EES), which will be used in the European Union (EU) to automatically monitor the border-crossings of third-country nationals. In such large-scale biometric systems, it is of utmost importance to ensure high quality of biometric data, since poor quality can cause recognition errors. Proposed biometric sample quality assessment algorithms can be categorized into monolithic and factor-specific approaches \cite{schlett_face_2022}. Monolithic algorithms are designed to directly extract a unified quality score. In this context, deep learning-based methods, e.g., \cite{FingerDL,CRFIQA}, have been shown to achieve competitive performance. In contrast, factor-specific approaches extract quality vectors consisting of quality scores associated to capture- and subject-related components. Each component corresponds to a specific defect that is known to negatively affect the utility of the biometric sample. For example, overexposure or defocus represent capture-related defects that decrease the utility of a  face image such that the corresponding subject may not be correctly verified. The assessment of distinct quality components facilitates \emph{explainability} and \emph{actionable feedback}. That is, decisions, e.g., the rejection due to insufficient quality, can be further explained to the data subjects or operators who can take according actions.   

Similar to the aforementioned concept of a factor-specific quality assessment, the German Federal Office for Information Security introduced the Open Source Face Image Quality (OFIQ) software. This software measures the quality of facial images and serves as a reference for algorithms compliant with the ISO/IEC DIS 29794-5 \cite{ISO-IEC-29794-5-DIS-FaceQuality-240129}. In addition to capture- and subject-related components, OFIQ also contains an algorithm that extracts a unified quality score.\footnote{Alternatively, a unified quality score might as well be derived from a set of component-based quality scores, as it is done in the current version of the NIST Fingerprint Image Quality algorithm (NFIQ 2) \cite{NIST-NFIQ2-FingerprintImageQuality-2021}.} Among other factors, image compression is an important quality component, since it may cause artefacts in face images. This is particularly relevant if the facial image was not captured by the biometric system itself, but is provided by the user or captured by external systems. For example, in some countries, e.g., Estonia, it is possible for the applicant of an identity card to digitally transmit the biometric face image \cite{estland_personal}. In such a scenario, the methods used to capture and store the image are unknown and therefore also the level of compression.  In the EES, biometric data is captured within the biometric systems of the member states and transmitted to the central system of the EU \cite{ees}. Despite the existing regulations \cite{council_of_european_union_commission_2019}, the use of many different systems implies that the quality of the transmitted data may vary from country to country. This means, transmitted face images must be checked for quality before storage. 

\begin{figure}[!t]
  \centering
  \subfigure[original]{
    \includegraphics[width=0.3025\linewidth]{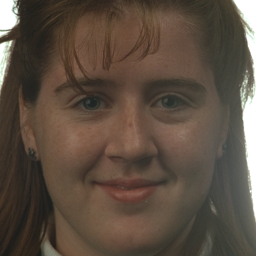}
    \label{fig:orig}
  }
  \subfigure[JPEG]{
    \includegraphics[width=0.3025\textwidth]{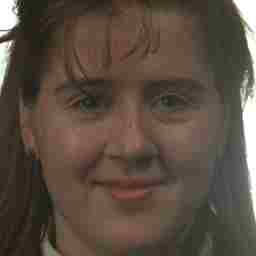}
    \label{fig:jpeg_2KB}
  }
  \subfigure[JPEG 2000]{
    \includegraphics[width=0.3025\textwidth]{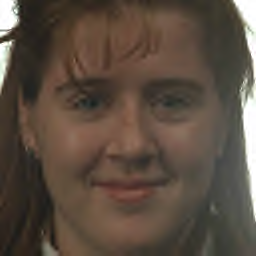}
    \label{fig:jp2_2KB}
  }\vspace{-0.2cm}
  \caption{Examples of image compression applied to an uncompressed face image (left) with JPEG (center) and JPEG 2000 (right) to a target file size of 2KB (image taken from FERET \cite{phillips_feret_1998}).}\vspace{-0.2cm}
  \label{fig:compression}
\end{figure}

Obviously, the size and format of an image file can be analysed to draw conclusions about the compression level. However, it is hardly possible to ensure that the file has not been converted to a different format or not compressed multiple times with different parameters. Moreover, image quality associated with a compression level also depends on the image content. The amount of high frequencies, i.e., fine details or rapid changes in intensity, may vary between facial images which leads to differences in the perceived quality at same compression rates. Consequently, a detection of compression artefacts based on the image content can help to uncover excessive application of image compression within a face recognition system.

The goal of this work is to investigate whether deep learning techniques are suitable for detecting compressed face images and determining the compression rate of facial images. In this context, the JPEG \cite{isoiec_jtc_1sc_29_coding_of_audio_picture_multimedia_and_hypermedia_information_isoiec_1994} and JPEG 2000 \cite{isoiec_jtc_1sc_29_coding_of_audio_picture_multimedia_and_hypermedia_information_isoiec_2019} image compression algorithms are considered, since these file formats are permitted for the storage of face images on passports and for the EES \cite{council_of_european_union_commission_2019}. Examples of heavily compressed face images are depicted as part of Figure~\ref{fig:compression}. 
In order to train neural networks, labels corresponding to the the level of compression are obtained by measuring the deviation of the compressed image from the original image. Two metrics are used for this comparison, i.e., the Peak Signal Noise Ratio (PSNR) and the  Structural Similarity (SSIM). The training of the neural networks is performed as a regression task to detect both, images compressed with JPEG as well as JPEG 2000. The resulting  networks output a value in a predefined range, which indicates the extent to which the image is affected by compression artefacts. This means, the presented scheme can be employed to detect image compression and to determine the severity of compression based on which highly compressed face images can be excluded from a face recognition pipeline to improve its biometric performance. For both application scenarios, experimental results confirm the soundness of the proposed approach.\footnote{A pre-trained model of the algorithm is made publicly available at: \url{https://github.com/BSI-OFIQ/OFIQ-Project}}  

The rest of this paper is organized as follows: previous works are briefly discussed in Section~\ref{sec:related}. The conducted data preparation is summarized in Section~\ref{sec:data}. The proposed methods are described in detail in Sect.~\ref{sec:method}. Subsequently, experimental results are reported and discussed in Section~\ref{sec:experiments}. In Section~\ref{sec:conclusion}, conclusions are drawn and potential future works are pointed out. 

\section{Related Works}\label{sec:related}
While the vast majority of research devoted towards the detection of image compression artefacts is aiming at the detection of image manipulation, e.g., in \cite{nikoukhah_jpeg_2019,zanardelli_image_2023}, or the removal of compression artefacts, e.g., in \cite{jiang_towards_2021}, only a few works explicitly deal with the detection of compression artefacts. Li et al. \cite{li_learning_2020} detect the compression of JPEG images by analysing the quantisation table in the JPEG header. This method has two main limitations: Only JPEG-compressed images can be analysed and, if an image is compressed multiple times, only the last compression can be tracked using the quantisation tables. Another work by Wang et al. \cite{wang_no-reference_2002} presents an algorithm that analyses  differences between neighbouring pixels and outputs a quality value. This algorithm can also only recognise JPEG compression. Uchida et al. \cite{uchida_pixelwise_2019} propose the use of a Convolutional Neural Network (CNN) for determining compression artefacts. Again, only JPEG compression can be recognised by this algorithm. It would be theoretically possible to re-train the approaches in \cite{wang_no-reference_2002,uchida_pixelwise_2019} to (additionally) detect JPEG 2000 compression artefacts. However, at the time of this writing, the source codes have not been made available by the authors for re-training the corresponding neural networks. Therefore, a comparison of the proposed systems with these works is left out in experiments. 

For the calculation of the compression rate, the current draft of the ISO/IEC  29794-5:2022 \cite{isoiec_jtc_1sc_37_biometrics_isoiec_2023} standard suggests dividing the actual file size by the uncompressed file size. The uncompressed file size is the product of the height, width, number of colour channels and number of bits per colour channel. The compression rate is then transformed into a quality score between 1 and 100. The problem with this method is that multiple conversions of the image could be applied. For example, an image could be converted to PNG format or bitmap format after being compressed to JPEG. In this case, the compression artefacts would remain the same, but the file size would increase. For this reason, such an algorithm is not useful for quality assessment.

\section{Data Preparation}\label{sec:data}

\subsection{Source Image Selection}
Artefact-free images of sufficient resolution are required as source images for the compression. This means that the images must not have been compressed with lossy compression resulting in visually noticeable artefacts. 
To this end, a total number of 48,000 face images were chosen from the Color FERET  \cite{phillips_feret_1998} and the FRGCv2 \cite{phillips_overview_2005} databases. While the Color FERET database contains images in lossless format (PNG), the FRGC database comprises artefact-free images. Since the FERET database only contains images with uniform background, only images with non-uniform backgrounds were selected from the FRGCv2 database. This should help the network to detect compression in facial images regardless of the background. In addition, 400 artefact-free images with good illumination were chosen from the Flickr Faces HQ \cite{8977347} database. 

\subsection{Pre-processing}
The images of the Color FERET and FRGCv2 were pre-processed by applying an alignment. For this purpose the SSD face detector\footnote{\url{https://github.com/sr6033/face-detection-with-OpenCV-and-DNN}} and the ADNet \cite{huang_adnet_2021} landmark estimation are used. Based on the detected landmarks a distortion-free affine transformation consisting of rotation, translation and scaling is computed that (approximately) maps the current position of the eyes, nose tip and mouth corners to pre-defined target positions. When applying this affine transformation, the image is cropped to the desired dimensions. Pre-processed images exhibit an inter-eye distance of approximately 260 pixels.

\subsection{Data Augmentation and Compression}

In order to obtain different compression variants of the images, two programs were selected for JPEG and JPEG 2000 compression: IrfanView \cite{irfanview} and ImageMagick \cite{imagemagick}. IrfanView is a proprietary software, while ImageMagick is an open source software. Note that the results of the two programs generally differ for same quality parameters, in particular for JPEG 2000 compression.

Training and validation sets are created based on the FERET and FRGCv2 databases and a test set is created based the FFHQ database. Prior to applying compression, various degrees of scaling and rotation are applied to some of the images. This is done to ensure that the compression artefacts visible in the images are of varying sizes and not always parallel to the image axes.

Each aligned image is scaled to an randomized IED. That is, the respective IED is randomly chosen from the set $\{60, 70, \dots, 130, 140, 200\}$. With 50\% probability, the scaled images are then rotated around the image centre by an angle randomly chosen from the integer interval $[-8, 8]$. All scaled and potentially rotated images are  compressed multiple times:
\begin{itemize}
    \item As JPEG using ImageMagick with every even compression quality value in the integer range $[20,100]$.
    \item As JPEG2000 using IrfanView  with every even compression quality value in the integer range $[20,100]$.
    \item As JPEG2000 using ImageMagick with every odd compression quality value in the integer range $[31, 99]$. 
\end{itemize}

For images that had been rotated before compression, the compressed images are rotated back by the negative angle around the image centre. The uncompressed  images are augmented by horizontal flipping. Finally, all images are scaled and cropped to 248$\times$248 pixels showing the inner face region. This results in a training set of approximately 3 million compressed images and 96,000 uncompressed images. For performing the training of a neural network as classification task, this training set would be highly unbalanced. However, in this work the training of  neural networks is performed as regression task with the aim of estimating the degree of applied image compression. This means, the uncompressed images represent the special case in which no compression is applied. Nonetheless, with  appropriate decision thresholds, trained networks can also be used to merely detect the presence of image compression artefacts.

For compilation of the test sets, the source face images of the Flickr Faces HQ database are processed in a similar manner to obtain two test sets referred to as Flickr-Rotated and Flickr-Upright. 

For the Flickr-Rotated set, each image is scaled to an inter-eye-distance randomly chosen from the set $\{60,90,120,140\}$. With 50\% probability, the scaled images are then rotated around the image centre by an angle randomly chosen from the integer interval $[-15, 15]$. All scaled and potentially rotated images are then compressed using IrfanView as follows: 
\begin{itemize}
    \item As JPEG with a random compression quality chosen from the set \newline$\{20,30,40,50,60,70\}$. 
    \item As JPEG2000 with a random compression quality chosen from the set \\$\{20,30,40,50,60,70\}$. 
\end{itemize}
For images that had been rotated before compression, the compressed images are rotated back by the negative angle around the image centre. 

For a Flickr-Upright set, the source images were processed in a similar way as for the Flickr-Rotated set but without applying rotations. For both sets the uncompressed images are augmented by horizontal flipping. As for the training set, in a final step, images are scaled and cropped to 248$\times$248 pixels. Both test sets consist of 800 uncompressed images, 800 images compressed with JPEG and 800 images compressed with JPEG 2000, i.e., 1,600 compressed images in total.



\subsection{Label Creation}
Depending on whether compression has been applied or not, the labels “compressed” and “uncompressed” are assigned, respectively.  
Further, two reference-based image quality assessment methods were used to create numeric labels for training the networks, i.e., PSNR and SSIM. The PSNR and SSIM values of compressed images were calculated for each image based on as the difference between the image and the corresponding uncompressed image (i.e., the source image). A min-max normalisation was then carried out for the PSNR values for mapping to the value range \([0,1]\). The SSIM values do not need to be normalised as they are already exclusively in the value range \([0,1]\). Uncompressed images were assigned to the label value 1.

\section{Proposed Method}\label{sec:method}

\subsection{Network Selection}
EfficientNetV2-B0 was selected as the network, since it provides a good trade-off between accuracy and resource consumption. This network achieves high efficiency which is crucial for operational deployments. This is particularly the case for OFIQ in which numerous quality components have to be estimated.  EfficientNetV2-B0 is the smallest variant of the EfficientNetV2 family. These were pre-trained on ImageNet for object classification. The EfficientNetV2 networks are characterised by the fact that they offer good precision at high speed. The EfficientNetV2-B0 was trained on ImageNet with an output layer with 1,000 classes \cite{tan_efficientnetv2_2021}. To adapt the network to regression, the output layer was replaced by a linear layer with an output neuron. This neuron is connected to each neuron of the previous layer with a weight and outputs a real value. After training, this value should indicate how much the image has been compressed. 

A single network is trained to detect both compression artefacts resulting from JPEG as well as JPEG 2000 image compression. To this end, the entire training set consisting of face images compressed with JPEG, JPEG 2000 and uncompressed images are fed to a single network in the training stage. This procedure is applied with either PSNR or SSIM labels. The obtained score has a lower-is-worse semantics and can be mapped to a integer-based scalar value in the range [0,100] as required in \cite{ISO-IEC-29794-5-DIS-FaceQuality-240129}. This is done by employing a suitable sigmoid function adjusted to the score distribution observed during the training stage.

\setlength{\tabcolsep}{6pt}
\begin{table}[!t]
\center
\caption{Selected hyperparameters for the proposed approach.}\vspace{-0.2cm}
\begin{small}
\begin{tabular}{lc}
    \toprule
    \textbf{Hyperparameter} & \textbf{Optimal value} \\
    \midrule
    Training data used & all \\ 
    Number of epochs & 10 \\ 
    Trainable layers & all \\ 
    Batch size & 256 \\ 
    Learning rate & 0.001 \\ 
    Image resolution & 256$\times$256 \\ 
    \bottomrule
\end{tabular}\label{tab:hyperparams}\vspace{-0.2cm}
\end{small}
\end{table}

\subsection{Hyperparameter Optimisation}
 Hyperparameter optimisation is exclusively performed using PSNR labels. It is assumed that the hyperparameters found to be optimal for this setting will also be optimal for networks trained with the SSIM labels. Optimal hyperparameters are determined by performing an exhaustive search through a manually specified subset of the hyperparameter space. 
 
A split of 80\% training data and 20\% validation data was chosen. The following hyperparameters were selected for the optimisation: the size of the training set, number of epochs, number of trainable layers, batch size, learning rate and image resolution. Adam and the Mean Squared Error were used as the optimiser and error function, as these are frequently used for regression tasks. The selected hyperparameters are considered isolated and it is assumed that they impact the accuracy independently of each other, even though there may exist correlations between them. In cases where only slight differences in accuracy are recognisable between different hyperparameter values, the hyperparameters are selected with regard to the resulting time required for training.  

Table~\ref{tab:hyperparams} lists the hyperparameters found to be optimal. Best results were achieved for using the entire amount of training data, i.e., all 80\%. This means that the employed data augmentation is helpful even if the resulting training set is relatively large. On the contrary, the amount of required epochs is relatively low, even for training all layers which generally revealed best results. 

\section{Results}\label{sec:experiments}

Two models are trained with PSNR and SSIM labels referred to as $M_{\mathit{PSNR}}$ and $M_{\mathit{SSIM}}$, respectively.
In a first experiment, the classification task, i.e., compression detection, is considered. This means the score distributions of quality scores obtained for uncompressed and compressed images are calculated based on which Detection Error Trade-Off (DET) curves are plotted. Additionally, classification accuracy is reported in terms of Equal Error Rate (EER) as the single point where the false positive rate is equal to the false negative rate. Further, the F1 scores are reported using the decision threshold of the EER operation point. As mentioned earlier, a comparison to previously published approaches is hampered due to the unavailability of source code. Theoretically, face image quality algorithms that extract unified quality scores could be compared against the proposed approach. However, in the detection task this is not meaningful: in practice, degradations caused by quality factors other than image compression are to be expected; that is, quality assessment algorithms that extract a unified quality score can not be used in practise to solely detect artefacts resulting from JPEG or JPEG 2000 image compression.   

\begin{table}[!t]
\center
\caption{Detection accuracy of trained models in terms of EER $\downarrow$ (in \%).}\vspace{-0.2cm}
\begin{small}
\begin{tabular}{lcccc}
    \toprule
    \textbf{Model } & \textbf{Test Set} & \textbf{JPEG} & \textbf{JPEG 2000} & \textbf{All}\\
    \midrule
    \multirow{2}{*}{$M_{\mathit{PSNR}}$} & Flickr-Upright & 0.875 & 2.125 & 2.0 \\ 
     & Flickr-Rotated  & 1.375 & 4.25 & 3.375 \\
     \midrule
    \multirow{2}{*}{$M_{\mathit{SSIM}}$} & Flickr-Upright & 3.0 & 4.625 & 3.375\\ 
     & Flickr-Rotated  & 4.5 & 6.5 & 5.875 \\ 
    \bottomrule
\end{tabular}\label{tab:results}
\end{small}
\end{table}

\begin{table}[!t]
\center
\caption{Detection accuracy of trained models in terms of F1 score $\uparrow$.}\vspace{-0.2cm}
\begin{small}
\begin{tabular}{lcccc}
    \toprule
    \textbf{Model } & \textbf{Test Set} & \textbf{JPEG} & \textbf{JPEG 2000} & \textbf{All}\\
    \midrule
    \multirow{2}{*}{$M_{\mathit{PSNR}}$} & Flickr-Upright & 0.9934 & 0.9840 & 0.9849 \\ 
     & Flickr-Rotated  & 0.9897 & 0.9678 & 0.9745 \\
     \midrule
    \multirow{2}{*}{$M_{\mathit{SSIM}}$} & Flickr-Upright & 0.9773 & 0.9647 & 0.9745\\ 
     & Flickr-Rotated  & 0.9659 & 0.9504 & 0.9553 \\ 
    \bottomrule
\end{tabular}\label{tab:resultsf1}\vspace{-0.2cm}
\end{small}
\end{table}

\begin{figure}[!t]
  \centering
  \subfigure[Flickr-Upright]{
    \includegraphics[width=0.475\linewidth]{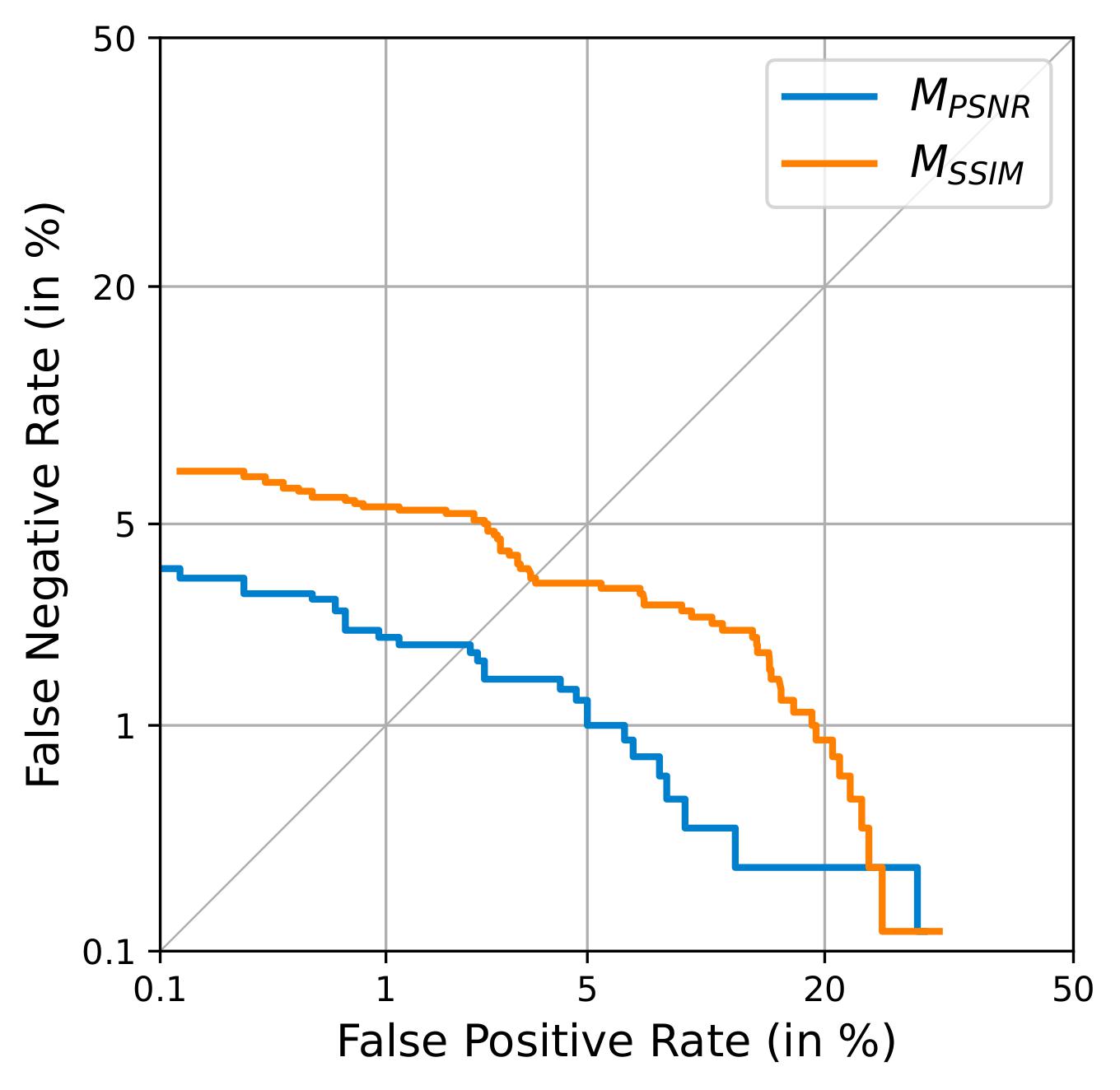}
  }
  \subfigure[Flickr-Rotated]{
    \includegraphics[width=0.475\textwidth]{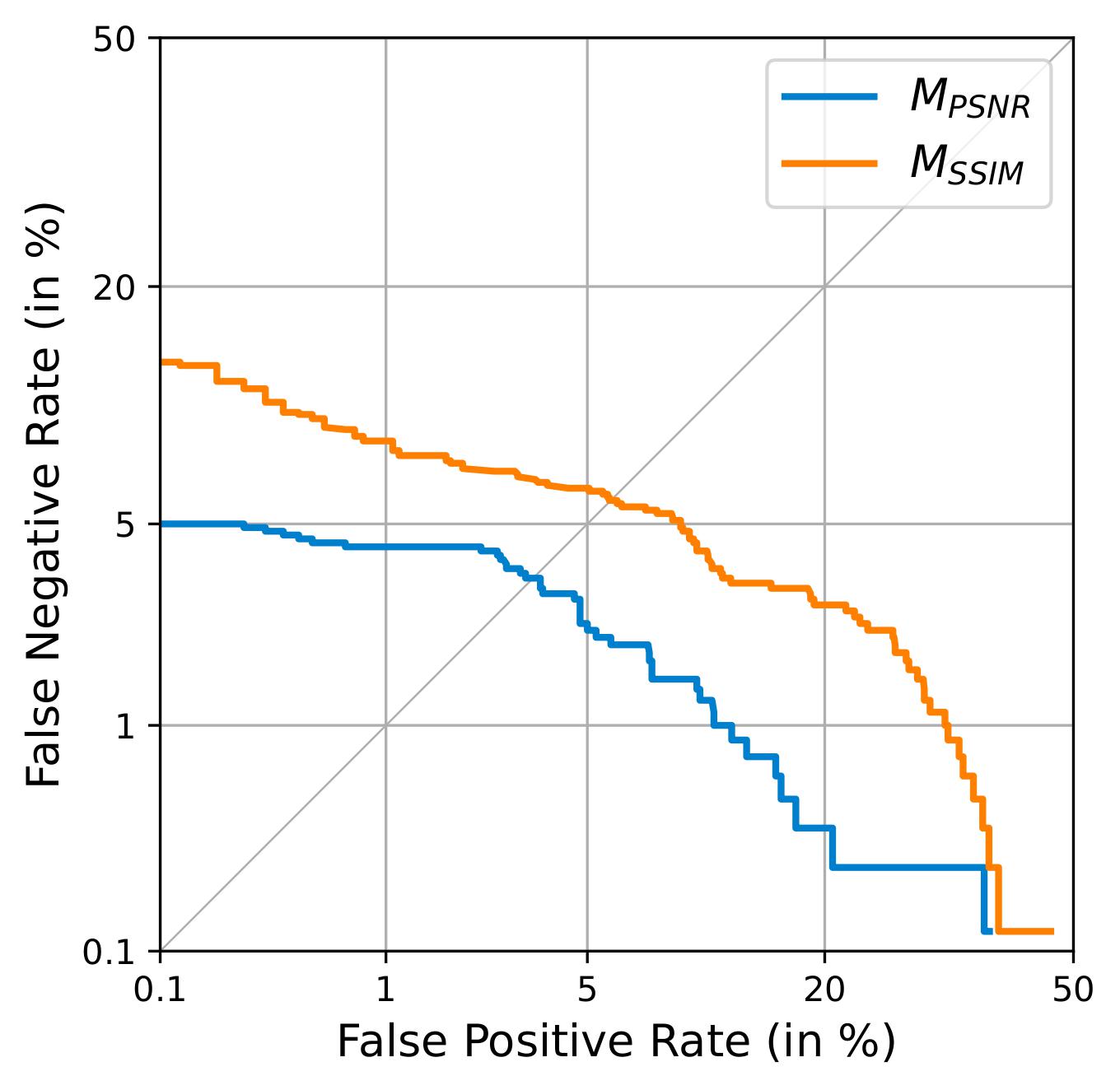}
  }\vspace{-0.2cm}
  \caption{DET curves for the detection networks on the different test sets.}
  \label{fig:single_dets}
\end{figure}

 The obtained results for both models on the different test sets are summarised in  Table~\ref{tab:results} and \ref{tab:resultsf1}. The corresponding DET curves are depicted in Figure~\ref{fig:single_dets}. As expected, detection error rates are generally smaller on the Flickr-Upright test set for both models. The $M_{\mathit{PSNR}}$ model outperforms the $M_{\mathit{SSIM}}$ model in terms of classification accuracy. Precisely, $M_{\mathit{PSNR}}$ model achieves an average EER of 2\% and on the Flickr-Upright test set and 3.375\% on the Flickr-Rotated test set. That is, compared to SSIM labels, PSNR labels appear to be better suitable for detecting the presence of image compression artefacts. Examples of predicted quality scores (scalar values) of the $M_{\mathit{PSNR}}$ model are shown in Table~\ref{tab:examples}. Further, it seems that JPEG compression artefacts are detected with higher accuracy than those caused by JPEG 2000 compression. This might be explained by the fact that at comparable compression rates JPEG 2000 is generally  causing less compression artefacts than JPEG compression. These observations are also reflected in the reported F1 scores.

\setlength{\tabcolsep}{3pt}
\begin{table}[!t]
\vspace{0.2cm}
\center
\caption{Examples of predicted quality scores by the $M_{\mathit{PSNR}}$ model.}\vspace{-0.2cm}
\begin{small}
\begin{tabular}{ccc}
 \textbf{\textsf{uncompressed}} & \textbf{\textsf{JPEG}} &  \textbf{\textsf{JPEG 2000}} \\
 \includegraphics[width=3.75cm]{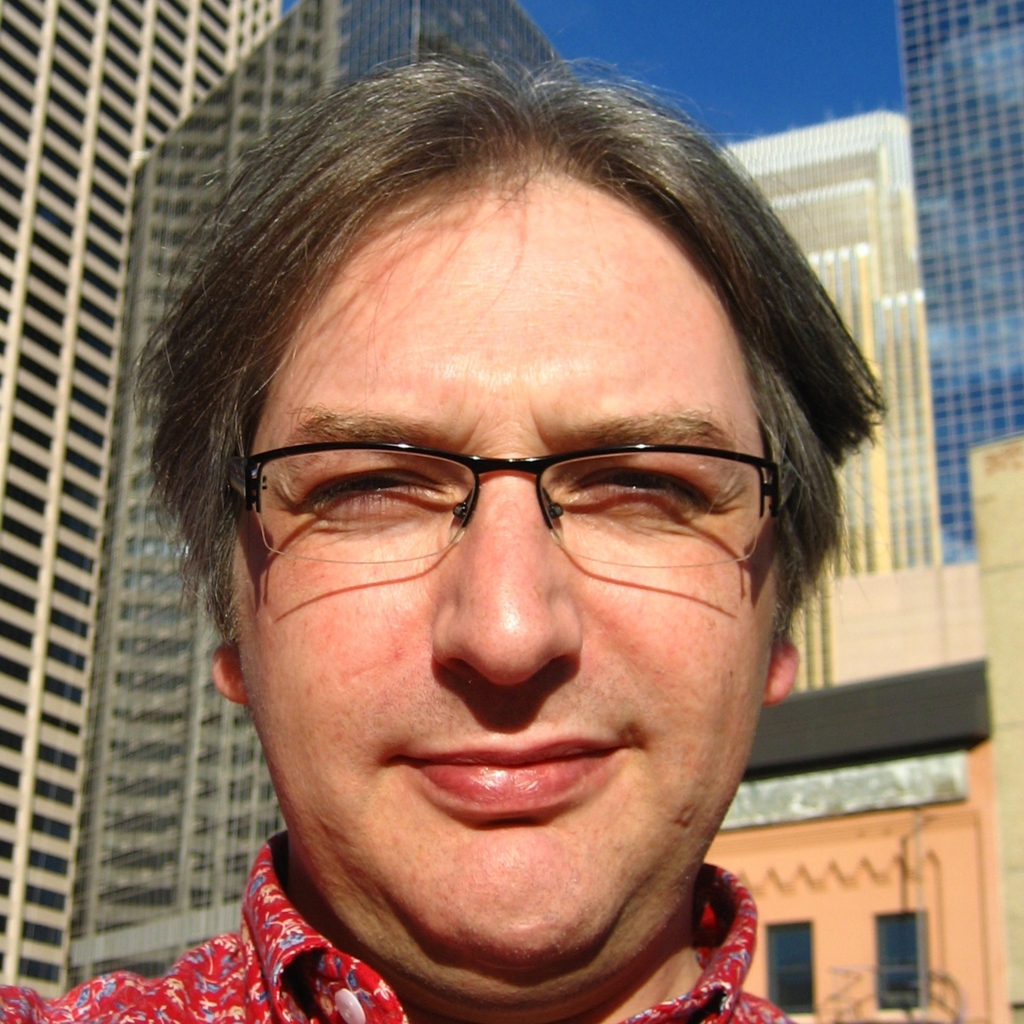} & \includegraphics[width=3.75cm]{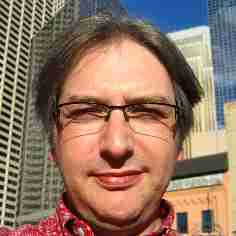} & \includegraphics[width=3.75cm]{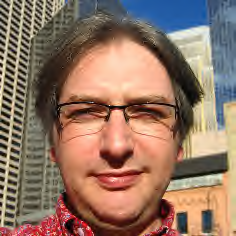}\\ 
  \textsf{quality: 98} & \textsf{quality: 2} & \textsf{quality: 0} 
\end{tabular}\label{tab:examples}\vspace{-0.2cm}
\end{small}
\end{table}

\setlength{\tabcolsep}{6pt}

In the second experiment, we evaluate whether the models are also able to accurately determine the strength of compression. The PSNR- or SSIM-based labels were only created for the image sets used during training. Therefore, 
the Spearman’s correlation coefficient $\rho$ between the network output and the employed compression quality parameter of JPEG and JPEG 2000 compression is calculated. For artefact-free face images a quality of 100 is assigned. 
Spearman’s correlation results can be found in Table~\ref{tab:correlation}. The $M_{\mathit{PSNR}}$ and $M_{\mathit{SSIM}}$ models exhibit a strong positive correlation with values around 0.9 for both test sets. However, the $M_{\mathit{SSIM}}$ shows slightly better results for predicting the strength of applied image compression.

\begin{table}[!t]
\center
\caption{Spearman’s correlation coefficients between network outputs and compression quality parameter.}\vspace{-0.2cm}
\begin{small}
\begin{tabular}{lcc}
    \toprule
    \textbf{Model } & \textbf{Test Set} &  $\mathbf{\rho} \uparrow$\\
    \midrule
   \multirow{2}{*}{$M_{\mathit{PSNR}}$} & Flickr-Upright &  0.9030 \\ 
     & Flickr-Rotated  & 0.8942 \\
     \midrule
    \multirow{2}{*}{$M_{\mathit{SSIM}}$} & Flickr-Upright &  0.9433\\ 
     & Flickr-Rotated  & 0.9269 \\
    \bottomrule\end{tabular}\label{tab:correlation}\vspace{-0.2cm}
\end{small}
\end{table}

Eventually, the usefulness of assessing the level of image compression during quality assessment in a face recognition system is analysed. For this purpose, a dataset that covers a wide range of quality defects should be employed. Since this requirement is typically fulfilled for in-the-wild face image datasets, these are frequently used for the evaluation of unified quality scores. Hence, 483,144 images of all 9,131 subjects of the VGGFace2 dataset \cite{VGGFace2}, which covers a large variety of quality levels and quality issues, are used. Performance is evaluated in terms of Error versus Discard Characteristic (EDC) curves. An EDC curve depends on an error type, i.e., False Non Match Rate (FNMR), a biometric recognition system, a set of comparisons each corresponding to a biometric sample pair, and a comparison score threshold corresponding to a starting error. The comparison score threshold associated with a starting error of FNMR=10\% is chosen. To compute an EDC curve, comparisons are progressively discarded based on the associated samples' lowest quality scores, and the error is computed for the remaining comparisons.   

The following face recognition systems are employed:
\begin{itemize}
\item ArcFace ResNet100 \cite{ArcFace}: an open-source face feature extractor based on the iResNet100 model and trained on MS1MV2. For face detection, RetinaFace was applied, using MediaPipe as fallback in case no face is detected. For this system the processing failed in 0.14\% of mated comparisons.
\item MagFace ResNet100 \cite{MagFace}: the open-source face feature extractor  based on the iResNet100 model and trained on MS1MV2 that can also be used to obtain a unified quality score. For face detection, SSD was used with RetinaFace as fallback if no face is detected. In 0.04\% of all mated comparisons, errors occurred.
\item Cognitec (Version 9.3.2.0): a commercial off-the-shelf face recognition system. In 3.07\% of the mated comparisons, errors occurred for this system when performing mated comparisons.
\item Paravision (Version 1.0.6): a commercial off-the-shelf face recognition system. This system failed to produce scores in 0.22\% of all mated comparisons.
\end{itemize}



\begin{figure}[thbp]
  \centering
  \subfigure[$M_{\mathit{PSNR}}$]{
    \includegraphics[width=0.475\linewidth]{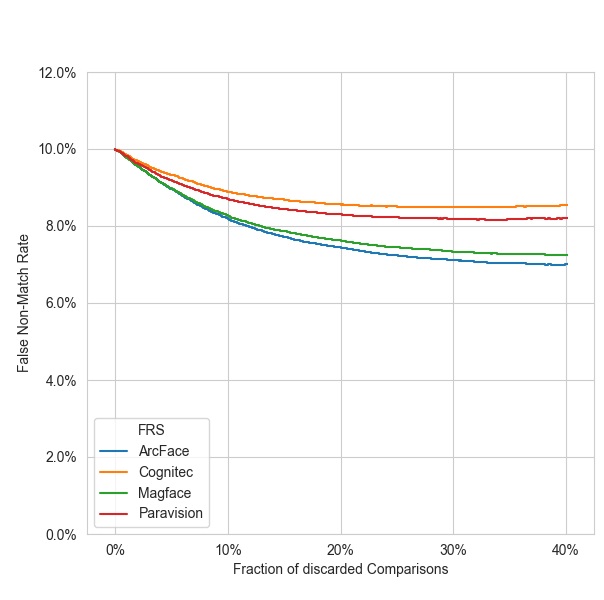}
  }
  \subfigure[$M_{\mathit{SSIM}}$]{
    \includegraphics[width=0.475\textwidth]{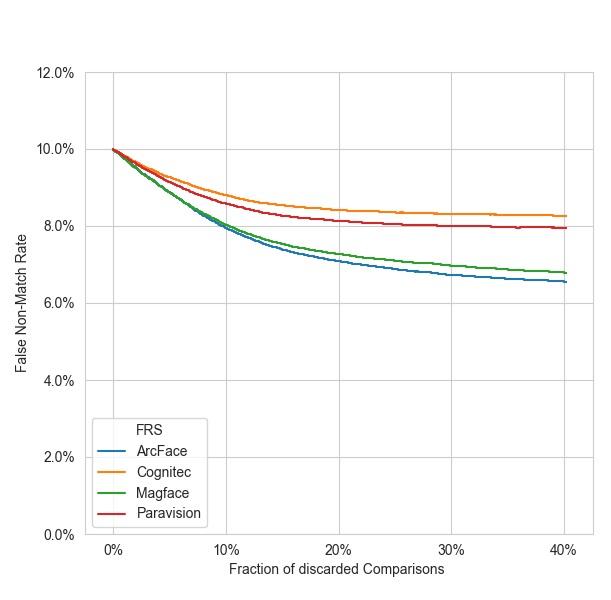}
  }\vspace{-0.2cm}
  \caption{EDC curves for different models and face recognition systems.}\vspace{-0.2cm}
  \label{fig:edc}
\end{figure}

Obtained EDC curves are shown in Figure~\ref{fig:edc}. For the two open-source face recognition systems a decrease of the FNMR down to approximately 8\%  is achieved for discarding 10\% of images that were determined to exhibit the worst quality in terms of image compression. For the commercial algorithms the reduction in FNMR is less pronounced. In general, the curves seem to flatten out after 15\% of the worst quality face images have been discarded. This indicates that face recognition accuracy is not affected by moderate image compression or that the amount of compressed images in the considered test set is limited. Nevertheless, the obtained results underline the importance of detecting severely compressed images in face recognition systems. Moreover, it is observable that the results are slightly better for the $M_{\mathit{SSIM}}$ model, as it was also reflected in the correlation analysis in Table~\ref{tab:correlation}. It is important to note that the obtained reduction in FNMR is not comparable to that achieved by algorithms designed to extract unified quality score (covering all degradation factors that have been present in a training dataset). For example, on the same test set, the quality scores derived from MagFace feature vectors achieve a reduction of the FNMR to about 6\% for discarding 10\% of the worst quality face images. However, a comparison against such algorithms is not fair and deliberately avoided. In addition, the quality assessment algorithms that aim for a unified quality assessment lack explainability, which is very important for operational deployments of facial recognition technologies.

\section{Conclusion and Future Work}\label{sec:conclusion}


The explainability of biometric systems has become increasingly important in the recent past, in particular due to the advances in deep learning-based algorithms. Such algorithms achieve high performance while their decisions are hardly explainable. In the case of operational applications, however, explainable decisions are required for users and operators in order to obtain usable feedback. In the development of the OFIQ software, which is targeted to establish a standard for face image quality assessment, this issue is addressed by measuring so-called quality components, e.g., defocus, brightness, or over-/under-exposure. Based on a set of quality components (defined in \cite{ISO-IEC-29794-5-DIS-FaceQuality-240129}), explainable decisions can be made and according actions can be taken by users or operators to eliminate detected defects. For measuring quality components, the use of deep learning-based methods is a viable solution. In this context, the detection of compression artefacts represents an integral part for the quality assessment of operational face recognition systems.  

This work investigated the feasibility of deep learning-based detection of compression artefacts in face images. It was shown that neural networks can be trained to reliably detect the presence of image compression, in particular JPEG and JPEG 2000 which are most relevant in applications of face recognition. Further, it was demonstrated that reference-based image quality assessment metrics, i.e., PSNR and SSIM, are suitable to obtain labels that were leveraged during the training of said networks. The resulting networks' outputs highly correlate with the strength of potentially employed compression (even in cases where images are scaled and/or rotated). This is highly useful for face image quality assessment and to filter out images exhibiting unacceptable quality with respect to image compression. The presented method is used in the publicly available OFIQ software for detecting compression artefacts as part of face image quality assessment.

In future work, two distinct networks could be trained to detect either JPEG or JPEG 2000 compression artefacts. 
The outputs of the two networks could subsequently be fused by returning the minimum of the final detection score, i.e., the lowest quality in terms of detected JPEG of JPEG 2000 compression artefacts. However, the detection score distributions produced by the two models may differ with respect to their range. Therefore, a proper normalisation would be required to be performed on a separate training set prior to the score fusion. Furthermore, note that the computational cost would increase in such an approach. 

\section*{Acknowledgements}
This work was supported by the Open Source Face Image Quality (OFIQ) project (P527 EESFM) of the German Federal Office for Information Security (BSI).

%
%
%
\bibliographystyle{splncs04}
%
\bibliography{lniguide}

\end{document}